\documentclass[conference]{IEEEtran}
\usepackage{filecontents}
\usepackage{lipsum}

\ifCLASSINFOpdf
\else
\fi
\usepackage{cite}
\usepackage{url}
\usepackage[pdftex]{graphicx}

\usepackage{bera}
\usepackage{listings}
\usepackage{xcolor}
\usepackage{esvect}
\usepackage{amsmath}
\usepackage{cleveref}

\colorlet{punct}{red!60!black}
\definecolor{background}{HTML}{FFFFFF}
\definecolor{delim}{RGB}{20,105,176}
\colorlet{numb}{magenta!60!black}

\lstdefinelanguage{json}{
    basicstyle=\normalfont\ttfamily,
    numbers=left,
    numberstyle=\scriptsize,
    stepnumber=1,
    numbersep=6pt,
    showstringspaces=false,
    breaklines=true,
    frame=lines,
    backgroundcolor=\color{background},
    literate=
     *{0}{{{\color{numb}0}}}{1}
      {1}{{{\color{numb}1}}}{1}
      {2}{{{\color{numb}2}}}{1}
      {3}{{{\color{numb}3}}}{1}
      {4}{{{\color{numb}4}}}{1}
      {5}{{{\color{numb}5}}}{1}
      {6}{{{\color{numb}6}}}{1}
      {7}{{{\color{numb}7}}}{1}
      {8}{{{\color{numb}8}}}{1}
      {9}{{{\color{numb}9}}}{1}
      {:}{{{\color{punct}{:}}}}{1}
      {,}{{{\color{punct}{,}}}}{1}
      {\{}{{{\color{delim}{\{}}}}{1}
      {\}}{{{\color{delim}{\}}}}}{1}
      {[}{{{\color{delim}{[}}}}{1}
      {]}{{{\color{delim}{]}}}}{1},
}

\begin{document}

%
\title{Web-based visualisation of head pose and facial expressions changes: monitoring human activity using depth data}

\author{\IEEEauthorblockN{Grigorios Kalliatakis}
\IEEEauthorblockA{School of Computer Science and\\Electronic Engineering\\
University of Essex, UK\\
Email: gkallia@essex.ac.uk}
\and
\IEEEauthorblockN{Nikolaos Vidakis}
\IEEEauthorblockA{Department of Informatics Engineering\\
Technological Educational\\ Institute of Crete, Greece\\
Email: nv@ie.teicrete.gr}
\and
\IEEEauthorblockN{Georgios Triantafyllidis}
\IEEEauthorblockA{Mediology Section, AD: MT\\
Aalborg University\\
Copenhagen, Denmark\\
Email: gt@create.aau.dk}}

\maketitle

\begin{abstract}
Despite significant recent advances in the field of head pose estimation and facial expression recognition, raising the cognitive level when analysing human activity presents serious challenges to current concepts. Motivated by the need of generating comprehensible visual representations from different sets of data, we introduce a system capable of monitoring human activity through head pose and facial expression changes, utilising an affordable 3D sensing technology (Microsoft Kinect sensor). An approach build on discriminative random regression forests was selected in order to rapidly and accurately estimate head pose changes in unconstrained environment. In order to complete the secondary process of recognising four universal dominant facial expressions (happiness, anger, sadness and surprise), emotion recognition via facial expressions (ERFE) was adopted. After that, a lightweight data exchange format (JavaScript Object Notation-JSON) is employed, in order to manipulate the data extracted from the two aforementioned settings. Such mechanism can yield a platform for objective and effortless assessment of human activity within the context of serious gaming and human-computer interaction.
\end{abstract}


%
\IEEEpeerreviewmaketitle

\section{Introduction}
Automatic and effective estimation of head pose is a challenging problem of computer vision systems. Since it is considered as a key element of human behaviour analysis, many applications would benefit from automatic and robust head pose estimation systems such as: (i) face recognition; (ii) human activity analysis; (iii) human-computer interaction and (iv) robotic vision. As a result, head pose estimation has drawn great attention from academia and a variety of techniques have been reported in the literature \cite{1,2,3,4}. 

Likewise, the field of facial expression analysis is still regarded as an enthusiastic issue in the latest research works \cite{6,7,8}. Due to its various purposes and applications, such as: (i) designing better human/machine interfaces; (ii) video gaming; (iii) computer generated animations and (iv) identification, facial expression analysis plays a key role in emotion recognition and thus contributes to the development of human-computer interaction systems \cite{5}.

With the recent technological advancements in the area of depth sensors, it is now feasible to perform large-scale data collection for subsequent analysis. Providing an objective assessment and evaluation, findings such as head pose changes and facial expression variations, can lead to valuable conclusions regarding the overall experience of users in many applications. One such application is the evaluation of the player\textquotesingle s training and ludology\footnote{ A borrowing from Latin word "ludus" (game), combined with an English element; The term has historically been used to describe the study of games.}  experience in the case of serious games such as \cite{vidakis2015,vidakis2014}.
\begin{figure}[!t]
\centering
\includegraphics[width=3.5in]{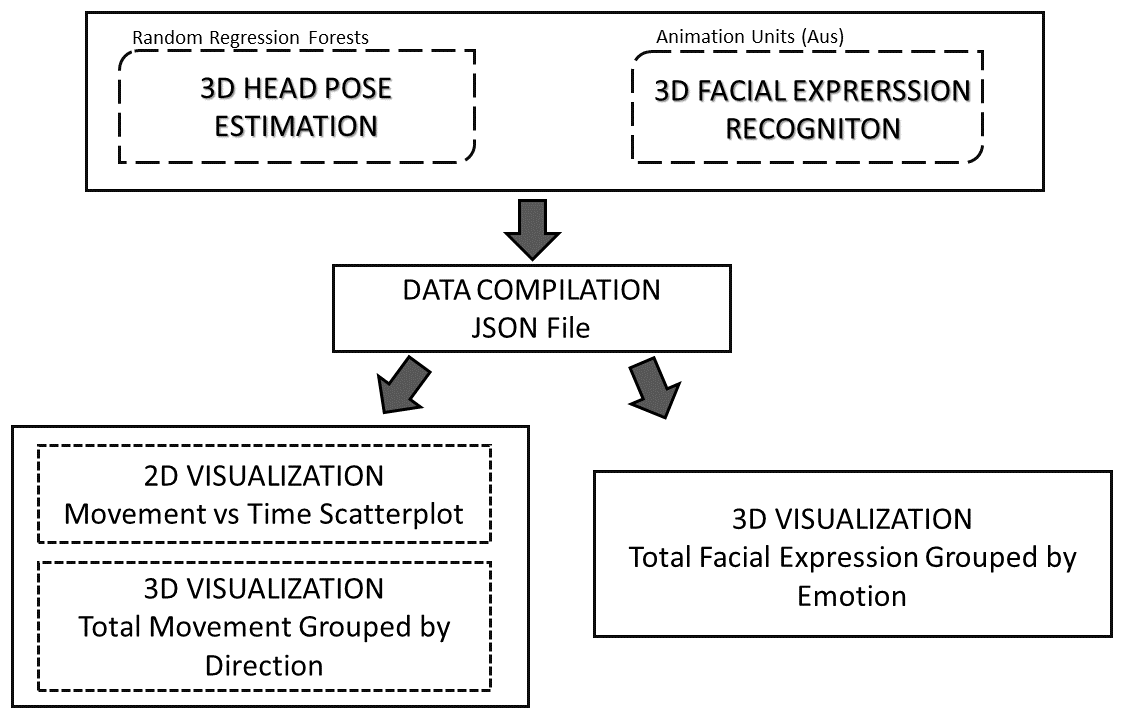}
\caption{An overview of the proposed system.}
\label{fig_sim}
\end{figure}
In this context, accessible visualisations can play a major part in that kind of assessment by creating encodings of data into visual channels that people can view and understand comfortably. The process of data visualisation is suitable for externalizing the facts and enabling people to understand and manipulate the results at a higher level. Additionally, visualisations can be used in several distinct ways to help tame the scale and complexity of the data so that it can be interpreted effortlessly.

In this paper, we address the problem of human activity monitoring through head pose and facial expressions changes and, in the same time, we introduce two innovative web-based visualisations for evaluating purposes with respect to those data. The proposed system consists of three distinctive components as shown in Fig.\ref{fig_sim}. First the real-time head pose estimation and facial expression events are separately obtained for different users sitting and moving their head without restriction in front of a Microsoft Kinect sensor for specified intervals. Experimental results on 20 different users show that the proposed system can achieve 83.95\% accuracy for head pose changes and 76.58\% accuracy for facial expressions recognition when validated against manually constructed ground truth data. Then the data for every user session are stored in a JSON file for offline manipulation. In the last step, two different types of visualisations are exploited: (i) a scatter plot for demonstrating head pose changes and intensities; (ii) 3D columns for presenting the players movement grouped by direction and the players facial expressions grouped by the dominant emotion respectively. As noted in our previous work \cite{Kalliatakis}, which was limited to head pose changes, \textit{the principal objective of this work is to acquire efficient and user-friendly visualisations in order to improve the understanding and the analysis of the captured data}, easily accessed through a web-page. 

The remainder of the paper is structured as follows: Section II describes the head pose estimation framework that was employed in our approach, while Section III explains the method for real-time emotion recognition via facial expressions. The data compilation phase is presented in Section IV, while Section V contains a brief description of the libraries that were used for presenting the data on the web, before presenting the actual web-based visualisations which were made for user activity assessment. Finally Section VI consists of a summary and concluding remarks.

\section{Head Pose Estimation Framework}
Systems relying on 3D data have demonstrated very good results for the task of head pose estimation, compared to 2D systems that have to overcome ambiguity in real time applications. We partly followed the approach of Fanelli \textit{et al.} \cite{9}, as it is regarded to be suitable for real time 3D head pose estimation, considering its robustness to the poor signal-to-noise ratio of current consumer depth cameras like Microsoft Kinect sensor. 

For this reason, regression forests are being extended in such a manner that depth patches belonging to a head can be discriminated and solely used for the prediction of the pose resulting in solving both the classification and regression problems respectively. While several works in the literature contemplate the case where the head is the only object present in the field of view \cite{10}, the proposed method concerns depth images where other parts of the body might be visible at the same time, and therefore need to be disjointed into image patches either belonging to the head or not. The system is able to perform on a frame-by-frame basis while it runs in real time without the need of initialization.
\begin{figure}[!t]
\centering
\includegraphics[width=3in]{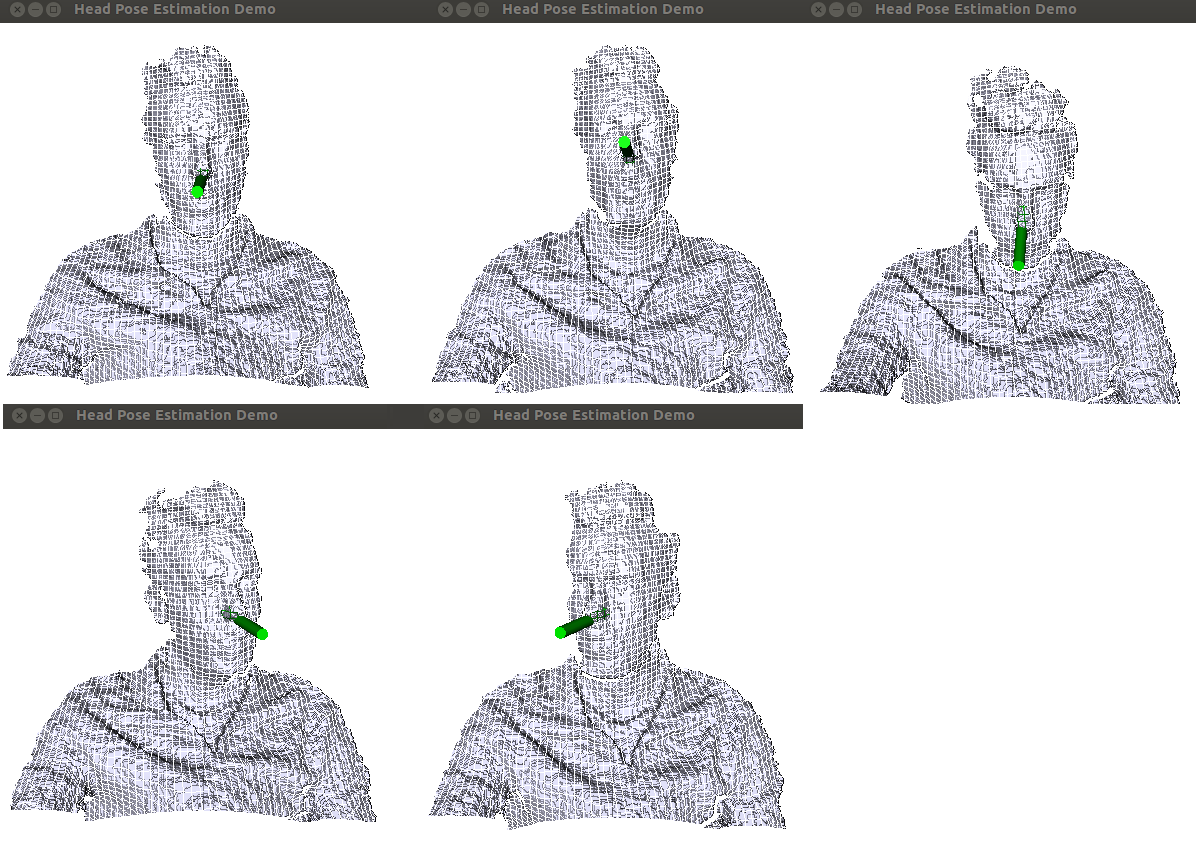}
\caption{Head pose estimation results.}
\label{head_pose_results}
\end{figure}
Forests of randomly trained trees are less sensitive to over-fitting and generalize better than decision trees independently. In the proposed setup \cite{11}, depth patches are annotated with class label and a vector containing the offset between the 3D points falling on the patch\textquotesingle s center and the head center location, plus the Euler rotation angles describing the head orientation. Randomness is imported in the training process, either in the set of training examples provided to each tree or in the set of tests used for optimization at each node, or even in both. When the pair of classification and regression are engaged, the aggregation of trees which simultaneously separate test data into positive cases (they represent part of the object of interest) are labeled as Discriminative Random Regression Forests (DRRF). This signifies that an extracted patch from a depth image is sent through all trees in the forest. The patch is evaluated at each node according to the stored binary test and passed either to the right or left child until a leaf node is reached \cite{12}, at which point it is classified and only if this classification outcome is positive (head leaf), a Gaussian distribution is recaptured and then used for casting a vote in a multidimensional continuous space which is stored at the leaf. Fig.\ref{head_pose_results} shows some processed frames regarding two DOF (\textit{pitch} and \textit{yaw}). Starting from left to right, the first row estimations displayed are: \textit{still}, \textit{up}, \textit{down}. The second row estimations are \textit{left} and \textit{right} correspondingly. All calculations derived from the difference between the exact previous frame and the current frame, at each iteration of the program. The green cylinder encodes both the estimated head center and direction of the face. 
\section{Facial Expression Recognition Framework}
\begin{figure}[!t]
\centering
\includegraphics[width=3in]{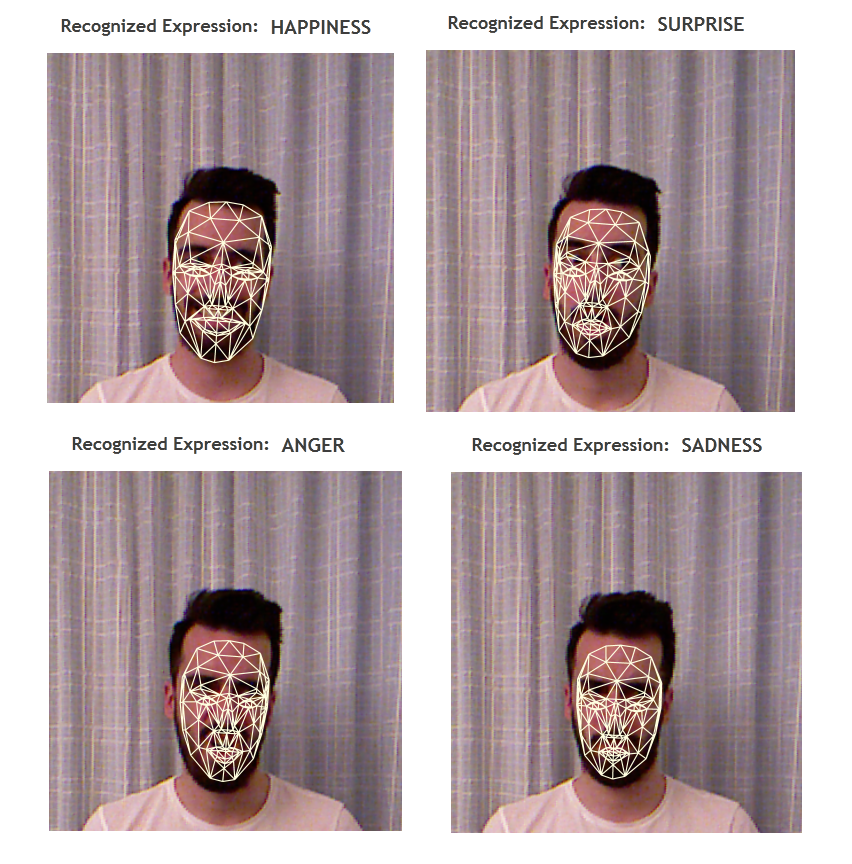}
\caption{Facial expression recognition (FER) results.}
\label{FER_results}
\end{figure}
Emotion recognition via facial expressions (ERFE) is a growing active research field in computer vision compared to other emotion channels, such as body actions and speech, primarily because superior expressive force and a larger application space is provided. 

A similar to \cite{13} approach was followed for real-time emotion recognition. Video sequences acquired from the Kinect sensor are regarded as input. Then face detection and feature extraction are performed on each frame of the stream. The Face Tracking SDK \cite{14}, which is included in Kinect\textquotesingle s Windows Developer toolkit, is used for tracking human faces with RGB and depth data captured from the sensor. Furthermore, facial animation units and 3D positions of semantic facial feature points can be computed by the face tracking engine, which can lead to emotion recognition via facial expressions. 

Face tracking results are expressed in terms of weights of six animation units, which belong to a subset of what is defined in the Candide3 model \cite{15}. Each AU, that is deltas from the neutral shape, is expressed as a numeric weight varying between $-$1 and $+$1, and the neutral states of AUs are normally assigned to 0. The AU\textquotesingle s feature of each frame can be written in the form of a 6-element vector:

\begin{equation}
\bar{a} = (A_{1},A_{2},A_{3},A_{4},A_{5},A_{6})
\end{equation}

where A1, A2, A3, A4, A5, and A6 refer to the weights of \textit{lip raiser},  \textit{jaw lower}, \textit{lip stretcher}, \textit{brow lower}, \textit{lip corner depressor}, and \textit{brow raiser}, respectively. For the purpose of this paper four different emotions were tested: \textit{anger}, \textit{happiness}, \textit{sadness} and \textit{surprise} as shown in Fig.\ref{FER_results}.

\section{Data Compilation}
In this section, the data compilation process based on the aforementioned frameworks is presented. Given the pitch $pitch_{t}$ and yaw $yaw_{t}$ intensities of the ongoing streaming frame, and the exact previous frame\textquotesingle s pitch $pitch_{t-1}$ and yaw $yaw_{t-1}$ intensities, the system operates in three steps as follows: (a) the differences regarding pitch and yaw are calculated by (2)\textendash(3);
\begin{equation}
pitchDiff = pitch_{t-1} - pitch_{t}
\end{equation}
\begin{equation}
yawDiff = yaw_{t-1} - yaw_{t}
\end{equation}
(b) then a threshold value was experimentally set around 4 in order for our system to ignore negligible head movements in all four directions tested; (c) finally, the changes with respect to the four different directions are given by (4)\textendash(7). 
\begin{equation}
up = pitchDiff > THRESH
\end{equation}
\begin{equation}
down = pitchDiff < THRESH
\end{equation}
\begin{equation}
left = yawDiff > THRESH
\end{equation}
\begin{equation}
right = yawDiff < THRESH
\end{equation}
Concerning the detection of emotions, boundaries for each Animation Unit had to be created in order to associate the vector obtained by the AU feature, as defined by (1), with the four main emotions. For example, (0.3, 0.1, 0.5, 0,$-$0.8, 0) corresponds to a happy face, which means showing teeth slightly, lip corner raised and stretched partly, and the brows are in the neutral position. We experimentally assembled the following equations (8)\textendash(11) for our test sessions:
\begin{equation}
sadness = A_{6}<0  \land  A_{5}> 0 
\end{equation}
\begin{equation}
surprise = (A_{2}<0.25  \lor  A_{2}> 0.25) \land A_{4}<0
\end{equation}
\begin{equation}
hapiness = A_{3}>0.4  \lor  A_{5}<0
\end{equation}
\begin{equation}
\begin{split}
anger = &((A_{4}>0 \land (A_{2}> 0.25 \lor A_{2}<-0.25) \\
             &\lor (A_{4}>0 \land A_{5}>0 ))
\end{split}
\end{equation}

\begin{figure}[!t]
\centering
\begin{lstlisting}[language=json,firstnumber=1]
[
   {"SessionDate": "2/Mar/16",
    "SessionData":[
                   {
                    "time": 2.23, 
                    "direction": "RIGHT",
                    "intensity": 6.78485
                   }
                  ]
\end{lstlisting}
\caption{JSON structure for head pose changes.}
\label{JSON_head_pose}
\end{figure}

\begin{figure}[!t]
\centering
\begin{lstlisting}[language=json,firstnumber=1]
[
   {"SessionDate": "10/Mar/16",
    "SessionData":[
                   {
                    "time": 7.98, 
                    "emotion": "ANGRY",
                   }
                  ]
\end{lstlisting}
\caption{JSON structure for facial expressions changes.}
\label{JSON_FER}
\end{figure}

Regarding the storage of the obtained data, JavaScript Object Notation (JSON) format was used mainly because of its lightweight nature, convenience in writing and reading and more importantly, as opposed to other formats such as XML, its suitability in generating and parsing tasks in various Ajax applications as described in \cite{16}. A record in an array was created for each user session, while an extra array was inside it, preserving three variables: \textit{time}, \textit{direction} and \textit{intensity} for each movement that was detected as shown in Fig.\ref{JSON_head_pose}. For facial expressions, a similar array was created, but in this case only two variables were required: \textit{time} and \textit{emotion}, as shown in Fig.\ref{JSON_FER}.

\section{Visualisation}
Many different approaches have been proposed in the literature to solve the problems of head pose estimation and facial expression recognition. However, very few focus on how those data must be presented in order to deliver a useful meaning conveniently. In this section, the final step of the proposed system is presented alongside the actual visualisation instances that can be found directly on the web. 
Furthermore, in the following subsections two JavaScript libraries, for data-driven document manipulation, are briefly exposed.

\subsection{D3: Data-driven Documents}
\textit{Data-Driven Documents} is a novel representation-transparent concept for web-based visualisations. This JavaScript library assists users at bringing data to life using varied technologies such as HTML for page content, CSS for aesthetics, JavaScript for interaction, SVG for vector graphics and so on. As claimed by Michael Bostock \textit{et al.} in \cite{17}, D3\textquotesingle s emphasis on web standards provides full capabilities of modern browsers while it combines powerful visualisation components and a data-driven approach to a shared representation of the page called the document object model (DOM).
However D3 must not be considered as a traditional visualisation framework because rather than introducing a novel graphical grammar, D3 solves the problem of efficiently manipulating documents based on data. Therefore D3\textquotesingle s fundamental contribution is a visualisation kernel, closer to other document transformers like \textit{jQuery} \cite{18} and \textit{CSS}, rather than a framework.

\subsection{Highcharts}
\textit{Highcharts} is a charting library written in pure JavaScript which suggests an easy way of adding interactive charts to web applications. Currently many different chart types are supported by this library such as box plot, pie charts, column charts etc. Many of these can be combined in one chart. This library was first released in late 2009, more details can be found in \cite{19}. One big advantage of this library lies in being packed with adapters, which means that it does not rely on one particular framework, but instead is pluggable to different frameworks. The default framework implementation of High-charts uses jQuery \cite{18}; hence the only requirement for users is to load the jQuery library before Highcharts. 

\subsection{Player Movement vs Time Graph }
\begin{figure}[!t]
\centering
\includegraphics[width=3.5in]{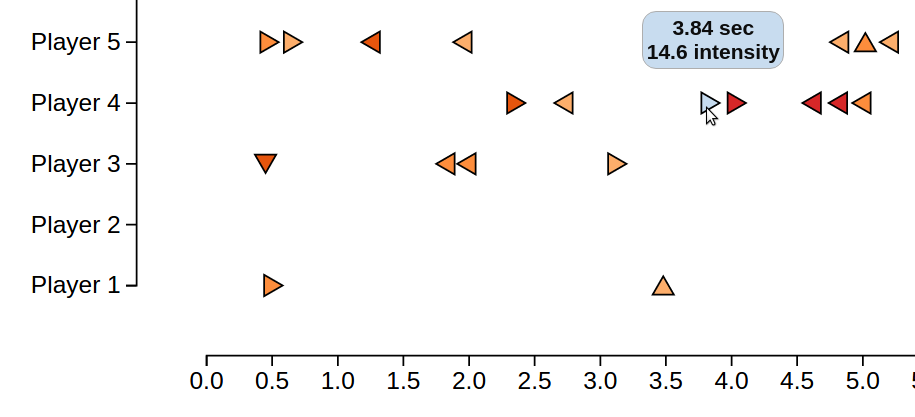}
\caption{2D Scatterplot of head pose changes.}
\label{scatterplot}
\end{figure}

In the final step of the proposed system, two visualisations were established for the
desirable web-based data interpretation regarding head pose changes. The first one is a 2D scatterplot displaying the head movement of the player over specified time period as shown in Fig.\ref{scatterplot}. 

In more details, x-axis represents the time scale in seconds during which the tests take place (Fig.\pageref{scatterplot} shows only a zoomed portion of the whole scatterplot graph), while each label in y-axis symbolizes each different user performing the test. Four different arrows imitate the movement of the human\textquotesingle s head in two DOF. Furthermore an additional feature is displayed when the mouse is hovering an arrow, showing the respective time each movement occurred and the intensity, which is based on how large the difference between the previous and the current frame was, as explained in Section IV. Apart from those elements, a color fluctuation is also evident which serves as an intensity indicator for each movement (the closer to red color the arrow is, the higher the intensity of the movement). One can easily examine the motion of the player that way, alongside its intensity, which adds a different dimension to the knowledge gained from the visualisation. 
The full version of this visualisation is available at: http://83.212.117.19/HeadPoseScatterplot/.

\subsection{Overall Movement Grouped by Direction}
\begin{figure}[!t]
\centering
\includegraphics[width=3.5in]{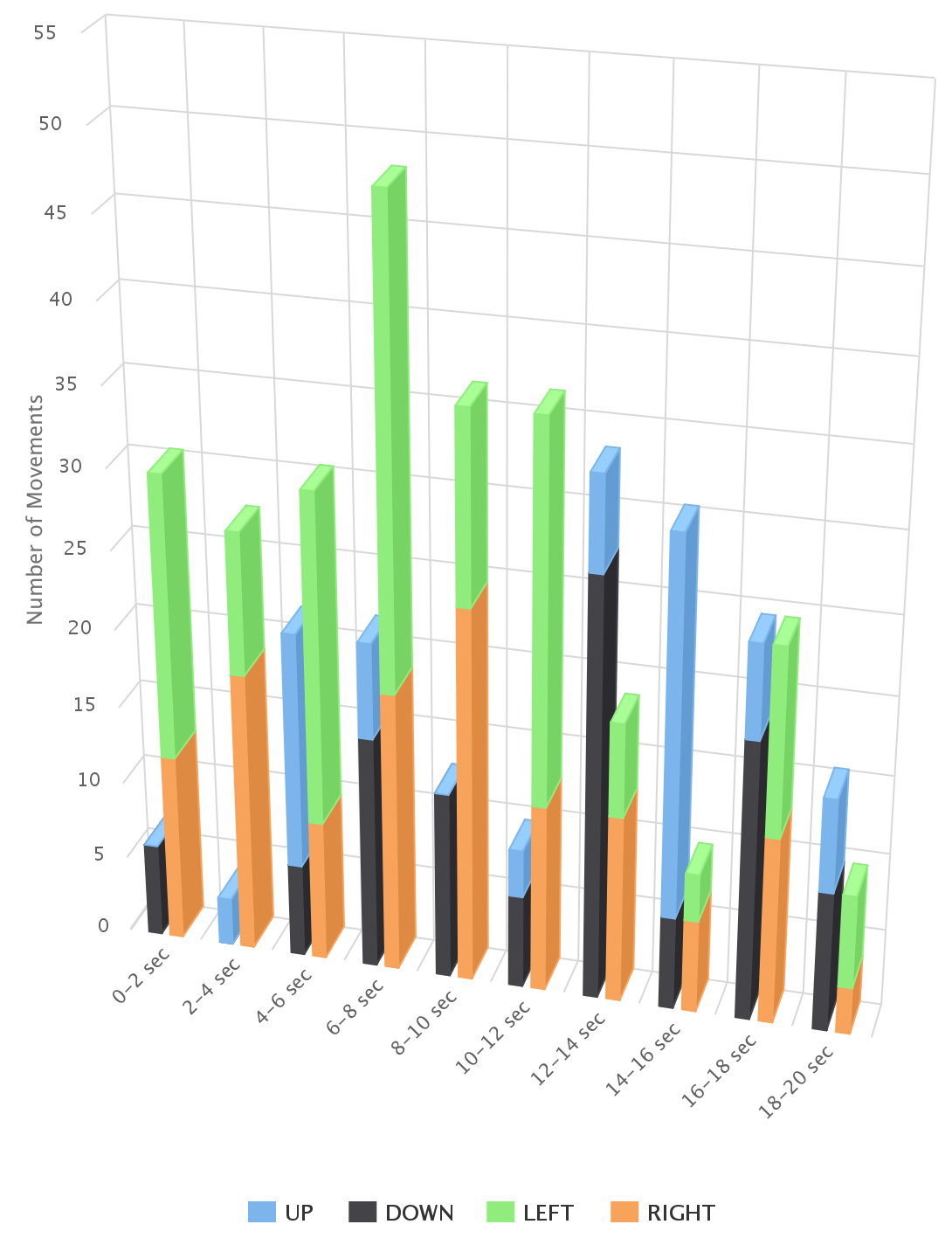}
\caption{ 3D column visualisation of head pose changes.}
\label{3D head pose changes}
\end{figure}

The second visualisation consists of a 3D column diagram which illustrates the aggregation of all head movements grouped by direction every two seconds as shown in Fig.\ref{3D head pose changes}. The four different directions are imitated by four different colors. In one hand, x-axis represents the time scale which is divided every two seconds until the end of the test. On the other hand, y-axis displays the number of movements for all the users that take part in the tests. Furthermore, when hovering above a column, the number of the corresponding direction summary is displayed. 
In this fashion, the dominant direction amongst all users every time interval is effortlessly assumed. Moreover, not so evenly distributed movements (e.g. columns between 2-4 seconds in Fig.\ref{3D head pose changes}) can lead into practical conclusions taking into account the nature of the test as well. The full version of the overall head movement visualisation is available at: http://83.212.117.19/HeadPose3D/. 

\subsection{Overall Facial Expressions Grouped by Emotion}
\begin{figure}[!h]
\centering
\includegraphics[width=3.5in]{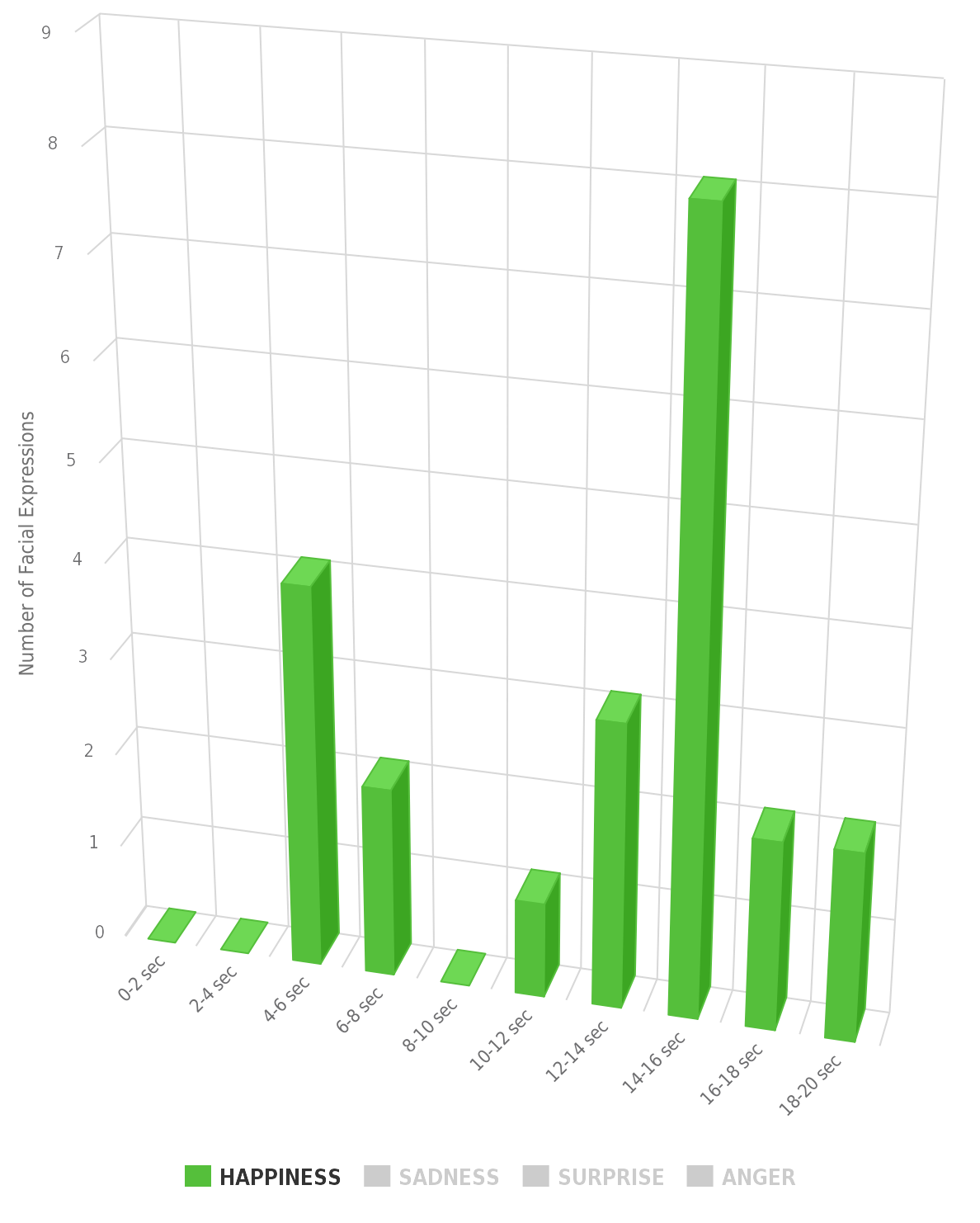}
\caption{3D column visualisation of "HAPPINESS".}
\label{happiness}
\end{figure}

The visualisation regarding the recognised emotions via facial expressions is assembled in the same fashion as the previous one. In this case the facial expressions are grouped by the recognised emotions. Fig.\ref{happiness} displays only one emotion, \textit{happiness}. However the rest of the recognised emotions can be set visible by clicking the corresponding check-box. The four different emotions are represented by four different colors. In one hand, x-axis represents the time scale which is divided every two seconds until the end of the test. On the other hand, y-axis displays the number of recognised emotions for all the users that take part in the tests. Furthermore, when hovering above a column, the number of the corresponding emotion summary is displayed. The full version of the overall facial expressions visualisation is available at: http://83.212.117.19/FacialExpression3D/.

\section{Conclusion}

In this paper we presented a system for generating efficient and user-friendly visualisations of head pose and facial expressions changes in the direction of human activity monitoring, utilising a consumer depth camera. Two selected approaches were extended in an applicable way for collecting and storing the required data using a key-value style lightweight exchanging format, JSON. Finally a 2D and two different 3D visualisations derived from the data compilation stage that can be easily accessed on the web. 
Our intention was to demonstrate that easily operated visualisations can provide a scaffold for objective and straightforward understanding of human activity across diverse applications, in the hope that it would provide a functional mechanism for future evaluations, and good baselines for human activity monitoring research. Interesting future directions will be to investigate whether other visualisation techniques can be more explanatory and effective, while fear and disgust could be included alongside the four main emotions.

\bibliographystyle{IEEEtran}
\bibliography{IEEEfull,testing} 

\end{document}